\documentclass[11 pt]{article}
\usepackage{graphics}
\usepackage{color}
\usepackage{graphicx}
\usepackage{amsmath,bbm,amssymb,mathrsfs, bm}
\usepackage{amsthm}
\usepackage{float} 
\usepackage{booktabs} 
\usepackage{array} 
\usepackage{verbatim} 
\usepackage{float} 
\usepackage{comment}
\usepackage{rotating}
\usepackage{lscape}
\usepackage{enumerate}
\usepackage{fullpage}
\usepackage{hyperref}
\usepackage{setspace}
\usepackage{multirow}
\usepackage{threeparttable}
\usepackage{algorithm}
\usepackage{algorithmic}

\usepackage[usenames,dvipsnames]{xcolor}
\usepackage[round]{natbib}
\usepackage{authblk}
\usepackage{enumitem}   

\theoremstyle{definition}

\usepackage{color}

\title{CLAIMED: A CLAssification-Incorporated Minimum Energy Design to explore a multivariate response surface with feasibility constraints}
\date{}

\author{Mert Y. Sengul*, Yao Song**, Linglin He**, Adri C.T. van Duin***, Ying Hung** and Tirthankar Dasgupta**}

\affil[*]{Department of Pharmaceutical Sciences, University of Maryland}
\affil[**]{Department of Statistics, Rutgers University}
\affil[***]{Department of Mechanical Engineering, Penn State University}

\begin{document}
	
\maketitle
	
\begin{abstract}
Motivated by the problem of optimization of force-field systems in physics using large-scale computer simulations, we consider exploration of a deterministic complex multivariate response surface. The objective is to find input combinations that generate output close to some desired or ``target'' vector. In spite of reducing the problem to exploration of the input space with respect to a one-dimensional loss function, the search is nontrivial and challenging due to infeasible input combinations, high dimensionalities of the input and output space and multiple ``desirable'' regions in the input space and the difficulty of emulating the objective function well with a surrogate model. We propose an approach that is based on combining machine learning techniques with smart experimental design ideas to locate multiple good regions in the input space.
\end{abstract}	
	
\section{Introduction} \label{sec:intro}
Exploration of multi-response physical/engineering systems with the objective of determining ``good'' points in the input space with desirable or target values of each output variable is typically a challenging problem. Such problems become even more complicated if the number of responses or outputs is large (possibly larger than the number of inputs), and certain combinations of inputs are ``infeasible'', in the sense they do not produce any reasonable output. Consider a system with $p$ input variables $X_1, \ldots, X_{p}$ and $q$ output variables $Y_1, \ldots, Y_{q}$. The problem is to determine ``good'' combinations of inputs $X_1, \ldots, X_{p}$ that produce responses $Y_1, \ldots, Y_{q}$ as close to some pre-defined ``target'' values $T_1, \ldots, T_{q}$ as possible. We assume that the functional relationships $Y_j = f_i(X_1, \ldots, X_p)$ for $j=1, \ldots, q$ are in principal known, as is the case in computer experiments, but they may be expensive to compute. We will denote the $q \times 1$ vectors of the responses and target values by $\mathbf{Y}$ and $\mathbf{T}$ respectively.

If $q=1$, i.e., for a single response, this problem can be formulated as a response surface exploration problem along the lines of \cite{BoxDraper1987}. When certain input combinations are infeasible, but such feasibility or infeasibility is not known before an experiment is conducted with that specific combination, the exploration is said to involve \emph{unknown constraints}. Such constraints need to be identified as the part of the exploration process. Such a problem has been addressed in engineering literature \cite{Henken2005}. When $q \ge 2$, a widely used approach is to reduce the response vector to some one dimensional loss function like the weighted squared error loss
\begin{eqnarray}
L(\mathbf{Y}, \mathbf{T}, \mathbf{W}) &=&  \sum_{j=1}^{q} \{ (Y_j - T_j)/w_j \}^2  \nonumber \\
&=& (\mathbf Y - \mathbf T)^{\text{T}} \mathbf{W}^{-1}  (\mathbf Y - \mathbf T), \label{eq:lossfn}
\end{eqnarray}
where $w_1, \ldots, w_{q}$ are a set of weights associated with the $q$ responses \cite{WuHamada2009} and $\mathbf{W}$ is a $q \times q$ diagonal matrix with entries $w_1, \ldots, w_q$. Such a method can also be applied with unknown constraints (e.g. \cite{Henken2007} where $q=2$). However, the loss function can still be a complex function of the inputs, making the exploration non-trivial.

\begin{figure}[ht]
\centering
\caption{Contour plot of loss function with two inputs and two outputs} \label{fig:contour}
\includegraphics[scale=.6]{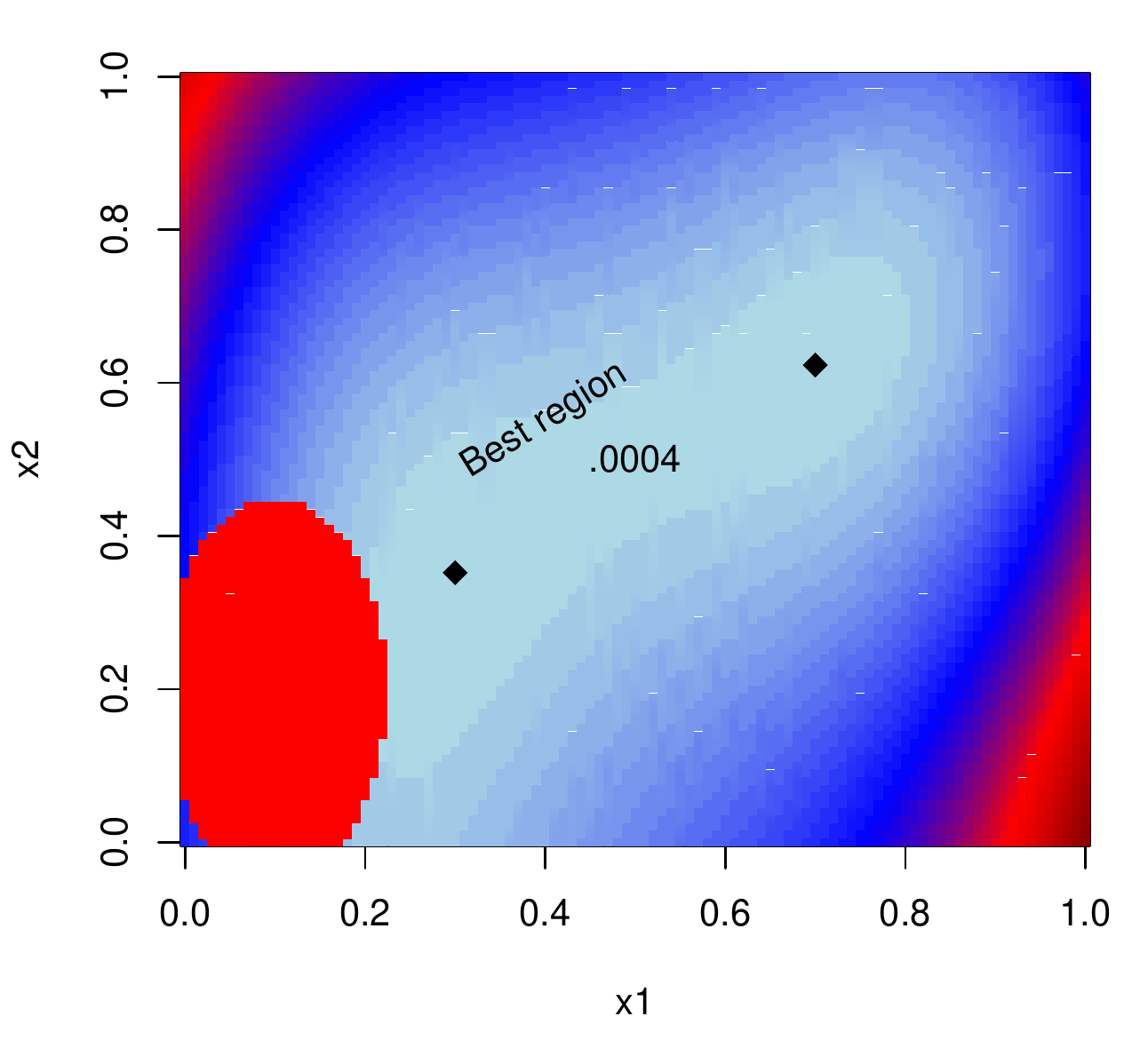}
\end{figure}

We illustrate this situation with a toy example with $p=q=2$. Let the input space be $[0,1]^2$, and suppose the input-output relations are described by the following equations:
$$ y_1 = \log (2+(x_1-0.7)(x_1-0.3)), \ y_2 = \log(2+x_2^2 + 0.5 x_2 - x_1),$$
that is, the first response only depends on $x_1$ whereas the second response depends on both $x_1$ and $x_2$. Let the target vector for $(y_1, y_2)$ be $(T_1, T_2) = \left( \log 2, \log 2 \right)$. Then for any weight vector $(w_1, w_2)$, the loss function $\sum_{j=1}^{2} \{ (y_j - T_j)/w_j \}^2$ is minimized at two points: $(0.3, (-0.5 + \sqrt{1.45})/2 )$ and $(0.7, (-0.5 + \sqrt{3.05})/2 )$ where it attains value zero.  Figure \ref{fig:contour} shows the contour plot of the loss function with weights $(1,20)$, in which the two optimum points are shown with diamond marks. However, as shown in the figure, loss function is virtually the same in the light blue band around these two optima - for example it is 0.0004 at $(0.5, 0.5)$. The red elliptical region near the bottom left corner shows the infeasible region, generated using the following logistic model:
\begin{eqnarray*}
u(x_1, x_2) &=& -0.25+\left( \frac{x_1-0.1}{0.25}\right)^2 + \left( \frac{x_2-0.2}{0.5}\right)^2, \\
\pi(u) &=& \exp(u)/\{1+\exp(u)\},
\end{eqnarray*}
and defining the binary feasibility variable $z = 0$ (infeasible) or 1 (feasible) according as $\pi < 0.5$ or $\pi \ge 0.5$.
	
\subsection{Motivating example: Optimization of ReaxFF systems} \label{sec:motivation}

Recent advances in materials science have established ReaxFF-based atomistic simulations as a promising method for atomistic level investigations in materials science. The ReaxFF is a force field that incorporates complex functions with associated inputs in order to describe the inter and intra-atomic interactions in materials systems. A typical ReaxFF force field consists of hundreds of parameters (inputs) per element type. During the development of a force field for a molecular system of interest, these parameters are optimized to reproduce reference values with reasonable accuracy. These reference values are molecular properties (e.g., bond lengths, bond angles, charges, and energies, etc.) of reference systems, also known as ``gold standards'', obtained by quantum chemistry methods (e.g., Density Functional Theory (DFT)) or experiments. In our notation defined in Section \ref{sec:intro}, for a ReaxFF system $X$'s represent parameters, $Y$'s represent the molecular properties and $T$'s represent the gold standards or target values. 

The conventional optimization method that is widely used by the force field developers is a sequential one input-at-at-time parabolic extrapolation method \cite{VanDuin2019}, which is not capable of switching between local minima to detect lowest loss regions in the input space. In contrast, this method is susceptible to being stuck in a local minimum, preventing parallelization of the optimization algorithm. Another big limitation of such one-factor-at-at-time approaches is its failure to capture important interactions among inputs. Due to these limitations in the conventional method, considerable effort has been directed toward finding a solution to this optimization problem using heuristic methods that include simulated annealing and genetic algorithm \cite{Iype2013, Larsson2013, Dittner2015} but none of them have been found efficient for exploring the parameter space comprehensively. 

The ReaxFF systems present several practical challenges to the exploration of complex input spaces in systems with multiple response. The true input output relationships are complex, typically involving systems of partial differential equations. Simulations for most ReaxFF systems are time consuming and expensive. The number of responses is large, making the loss function complex and multimodal. Another major complication arises from the fact that several parameter combinations do not produce meaningful outcomes, in the sense that either the simulation does not converge within the specified stopping time, or the results produce such a large discrepancy of the $Y$'s from the $T$'s that they are not meaningful at all. For many ReaxFF systems, the percentage of such infeasible combinations is larger than the percentage of feasible ones, making the process of output data generation even more expensive.

\subsection{Exploration vs Optimization} \label{ss:ExpvsOpt}

There exist several techniques and algorithms for global optimization of complex black box functions with and without constraints, Bayesian global optimization being the most popular among these \cite{Gramacy2016, Lobato2016, Muller2019}. Most of these algorithms extend the idea of the kriging-based expected improvement (EI) algorithm originally proposed by \cite{EI1998} and reviewed in \cite{Jones2001}, and are able to find the global optima of complex response surfaces quite efficiently.

However, our formulation of the problem is different from that in a typical global optimization problem, where the goal is to detect several ``good'' points that are spread within the feasible input space. To see why this is the case, assume that the contour in Figure \ref{fig:contour} represents the discrepancy function of a ReaxFF system associated with an alloy composed of nickel, chromium and molybdenum. The percentage of each metal type in the alloy determines its properties in real world applications. Clearly, one or two globally optimized parameter sets are unlikely to reproduce the properties of interest for all different alloy compositions, because each alloy requires specific properties that vary with parameter combinations. However, having several \emph{distinct} ``almost equally good'' points gives the flexibility to select one of these points that work best for a specific alloy composition. Consequently, if in addition to the two global optima we are able to locate several combinations of the ReaxFF parameters that are almost equally good but are located at a distance from each other, then the applicability of ReaxFF parameter sets significantly improves.

As demonstrated in \cite{MED2015}, BO-based methods such as the EI algorithm can usually identify global optima of complex response surfaces with high efficiency. However, when it comes to picking several points in ``good'' regions, it does not perform as well, because the points identified by BO algorithms tend to quickly cluster around the global optima. We will demonstrate this aspect of the EI algorithm in Section \ref{ss:mined}, and compare its performance with that of the proposed exploration strategy. We also show in our application with the Ni-Cr system in Section \ref{sec:NiCr}, how the exploration strategy helps us find several good parameter combinations, each of them being useful in a different way.

We propose an exploration strategy by developing a structured framework that aims at identifying points that are \emph{evenly spread} over the ``best'' regions where responses are reasonably close to their target values. The framework revolves around a smart experimental design strategy called the minimum energy design (MED) and supervised machine learning methods like classification. Unsupervised machine learning methods like clustering and simple data exploration and visualization techniques also form useful components of the framework. Such a combination, parallel to solving the engineering problem, facilitates important scientific understanding about the underlying process.

\subsection{Structure of the paper} \label{ss:structure}

In the next section, we describe the essential steps in the proposed framework assuming that the true input output relationships are known, or at least, can be simulated correctly whenever necessary. This assumption is true for all ReaxFF systems; however, in most cases the number of evaluations required to find a solution with the proposed approach may become prohibitive. In Section II, we also present some ideas to build surrogate models or ``emulators'' of the actual input-output functions, that can be used to overcome the problem of conducting a prohibitive number of expensive simulations. In Section III we demonstrate the effectiveness of the proposed approach using a few simulation studies. Finally, we demonstrate applications of the proposed approach using two actual examples of the Molybdenum Disulfide (MoS$_2$) ReaxFF system and Nickel-Chromium (Ni-Cr) binary ReaxFF system in Section IV, and present some conclusions and opportunities of future research in Section V.


\section{Essential steps involved in the proposed exploration approach} \label{sec:steps}

Space filling designs have found extensive use in efficient exploration of complex response surface and for identification of good regions in the input space. As explained in \cite{Joseph2016}, space filling designs evenly spread points all over experimental region with as few gaps or holes as possible. Therefore, ideally, if one can construct a very large space-filling design to cover the entire design space, then it is possible to identify the good input regions. However, as the dimension of the input space increases, the total number of points required to cover the entire space become prohibitively large, and additional strategies are required to facilitate exploration. Thus, most strategies for exploring large and complex input spaces rely on a sequential strategy in which an initial exploration is done using a standard space-filling design.

Our first step therefore is to create an initial space-filling design of $N$ points that does an initial exploration of the input space. Each of the $N$ design points in this initial design is a combination of the $p$ inputs. Several of these $N$ points are expected to be infeasible, i.e., they do not produce any output. Let $N_1$ and $N_0$ denote the number of feasible and infeasible points respectively, such that $N_1 + N_0= N$. The $N_1$ feasible points generate an $N_1 \times q$ matrix of responses, where each column corresponds to an output $Y$.

The second step is to fit a classification model using the information on $N_1$ feasible and $N_0$ infeasible points, that predicts the probability that a given combination of ReaxFF parameters will result in a feasible simulation output. The classification model can be parametric, e.g., logistic regression, or based on non-parametric or machine learning methods like random forests.

The third step is the most crucial in the exploration process and involves exploring the input space using the $N_1 \times q$ matrix of responses as the starting point, and finding the best regions with lowest loss making as few evaluations of new combinations as possible. This step will also incorporate the classification model fitted in the previous step. This will be done by an application of a recently proposed design strategy called minimum energy designs (MED) \cite{MED2015}.

As in the case of the ReaxFF problem, we assume that the true response functions are deterministic and known, but the feasibility region is unknown. However, with increase in the dimension of the input space $p$, the actual number of evaluations of the true function required to obtain good results may be prohibitive. In such cases, one may need to substitute the simulator by a cheap or fast surrogate called the ``emulator''. This is an optional fourth step in the optimization algorithm.

We now describe these four steps and illustrate each step using the toy example described in Section \ref{sec:intro} and used to generate Figure \ref{fig:contour}. 

\subsection{Initial design and data generation} \label{ss:initial design}

An initial space filling design is generated in which the levels of the $p$ input variables are simultaneously varied to produce $N$ different combinations. Simulations are conducted at each of these $N$ combinations and the outcome recorded. A $p$-dimensional input space can be explored using specific design strategies such as the Latin Hypercube sampling \cite{LHD1979}. The design generated using such a sampling strategy is called a Latin hypercube design (LHD). However, as observed by several authors, an arbitrary LHD does not necessarily have good space-filling properties and it is necessary to incorporate additional criteria like orthogonality of inputs and maximin distance (that maximizes the minimum distance between every pair of points in the design space). The initial design proposed for initial exploration of the input space is known as Orthogonal-maximin latin hypercube design (OMLHD) originally proposed in \cite{Hung2008}. The OMLHD algorithm can generate parameter combinations within ranges specific to each parameter that are multidimensionally uniformly distributed by reducing the pairwise correlation and maximizing the distance between parameters.

The outcome of each simulation obtained from the initial OMLHD is recorded as follows: (i) a binary outcome variable $Z$ taking values 0 and 1 according as whether the parameter combination is infeasible (does not produce any result) or feasible (produce results) and (ii) the values of the responses $Y_1, \ldots, Y_{q}$. The structure of the raw data matrix is shown in Table \ref{tab:data} in which the rows are arranged by values of $Z$ without loss of generality. The data would thus consist of $N$ rows and $p+q+1$ columns ($p$ input variables, $q$ responses, and one binary feasibility column). The response columns will be missing for the $N_0$ infeasible combinations, shown as rows $N_1+1, \ldots, N$ in Table \ref{tab:data}.

\begin{table}[htbp]
\small
\centering
\caption{Data generated from initial design} \label{tab:data}
\begin{tabular}{c | c c c | c c c | c}
\hline
No & \multicolumn{3}{c|}{Inputs} & \multicolumn{3}{c|}{Responses} & Feasibility \\ 
   &  $X_1$ & $\cdots$ & $X_p$ & $Y_1$ & $\cdots$ & $Y_q$ & $Z$ \\ \hline
1  &  $X_{11}$ & $\cdots$ & $X_{1p}$ & $Y_{11}$ & $\cdots$ & $Y_{1q}$ & 1 \\ 	
$\vdots$ & $\vdots$ & $\vdots$ & $\vdots$ & $\vdots$ & $\vdots$ & $\vdots$ & $\vdots$ \\
$N_1$ & $X_{N_1 1}$ & $\cdots$ & $X_{N_1 p}$ & $Y_{N_1 1}$ & $\cdots$ & $Y_{N_1 q}$ & 1 \\ 	\hline
$N_1 + 1$ & $X_{N_1+1, 1}$ & $\cdots$ & $X_{N_1+1, p}$ & $\cdots$ & $\cdots$ & $\cdots$ & 0 \\
$\vdots$ & $\vdots$ & $\vdots$ & $\vdots$ & $\vdots$ & $\vdots$ & $\vdots$ & $\vdots$ \\
$N$ & $X_{N 1}$ & $\cdots$ & $X_{N p}$ & $\cdots$ & $\cdots$ & $\cdots$ & 0 \\ \hline
\end{tabular}
\end{table}

\begin{figure}[ht]
\centering
\caption{Initial exploration of the loss function in the toy example using a 20-point Maximim LHD} \label{fig:initialtoy}
\includegraphics[scale=.6]{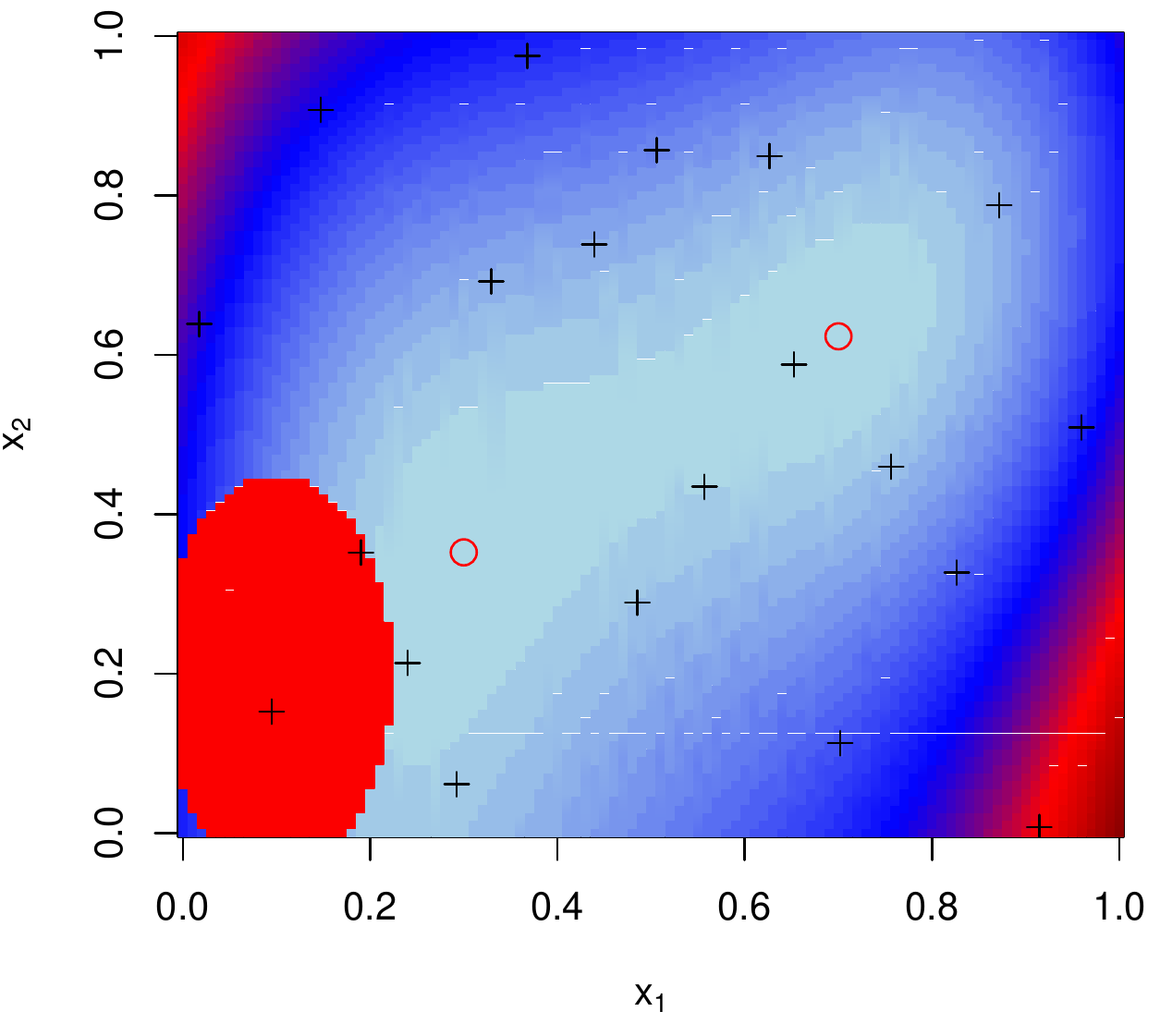} 
\end{figure}

Figure \ref{fig:initialtoy} demonstrates initial exploration of the loss function in the toy example described in Section \ref{sec:intro} using a 20-point maximin LHD. The design generates $N_1 = 18$ points (90\%) in the feasible region, which is consistent with the small size of the infeasible region. In our motivating example, this will typically not be the case, with the infeasible region often being larger than the feasible region making $N_0 > N_1$ in most cases.

In this example, the initial design also finds three points in the desired region, and one of them is close to one of the two optima. Again, this will rarely be the case, because as the dimension of the input space increases, the volume of the desired region will typically be very small compared to the total input space.

\subsection{Classification model} \label{ss:classification}

Having generated the data from the initial simulation experiment, the next task is to fit a classification model that predicts the feasibility of a combination of $X_1, \ldots, X_p$. Such a model is fit using the binary outcomes $Z$ applying a suitable supervised classification algorithm. There are several classification algorithms in classical statistics and machine learning \cite{StatisticalLearning2009}. Among these methods, logistic regression stands out as a versatile tool used in almost all scientific fields for several decades and is a natural candidate to be used. On the machine learning side, random forest classification \cite{Breiman2001} has attained tremendous popularity in scientific and engineering applications in the recent years. In practice, multiple algorithms can be tried and the one with best out-of-sample prediction performance can be chosen. Such a strategy necessitates splitting the data matrix into training and testing sets, fitting models using the training set and comparing them on the basis of certain performance metrics using the testing set. Two performance measures -- \emph{sensitivity} and \emph{specificity} -- can be considered for comparing the approaches and identifying the best one. Let TP, TN, FP and FN denote the numbers of true positives (correct identification of feasibility), true negatives (correct identification of non-feasibility), false positives (incorrect classification of a true infeasible point as feasible) and false negatives (incorrect classification of a true feasible point as infeasible). Then sensitivity is defined as the true positive rate measured as TP/(TP+FN), whereas specificity is defined as the true negative rate measured as TN/(FP+TN). 

\subsection{Finding points in the desired region using a Minimum Energy Algorithm} \label{ss:mined}

The primary objective of space-filling experimental designs is to spread points uniformly over the input space. The foregoing discussion in Section \ref{sec:intro} and Section \ref{ss:initial design} establishes that this is not the case in the current problem. We need a sequential design or active learning (\cite{Sung1995, Cohn1996}) strategy that helps avoid bad regions and generate more and more points from the desired region as the algorithm progresses. Thus the points sampled by the design should represent the response surface $r(x)$ under exploration, which means more points should be generated in regions where $r(x)$ is large and fewer points in regions where $r(x)$ is small. In our case, because regions of lower loss are desirable, the response surface $r(x)$ of interest can be taken as the inverse of the loss function $L(x)$ in (\ref{eq:lossfn}). Such an algorithm known as minimum energy designs (MED) was proposed by Joseph et al. (2015) \cite{MED2015}. We now briefly describe the minimum energy algorithm and explain its usage in the context of our problem again using the toy example.

Consider the problem of exploring a $p$-dimensional input space, where the range of each factor is scaled to $[0,1]$. Thus the design space is $[0,1]^p$. Let $q(x_i)$ be the weight associated with the $i$th design point $x_i$. Visualize $x_i$ as a charged particle in the box $[0,1]^p$ and $q(x_i)$ as the positive charge associated with it. Let $d(x_i, x_j)$ denote the Euclidean distance between the points $x_i$ and $x_j$. Then the design that minimizes the total potential energy for $n$ charged particles
$$ E = \sum_{i=1}^{n-1} \sum_{j=i+1}^n \frac{q(x_i) q(x_j)}{d(x_i, x_j)}, $$
is the MED. If the objective is to select samples that represent a positive response surface $r(x)$, the weight $q(x)$ should be chosen as $1/\left\{ r(x) \right\}^{1/(2p)}$.

Based on the above idea, Joseph et al. (2019) \cite{MED2019} proposed an efficient algorithm for computation of MED that optimizes a variant of the above energy function, and selects points to represent a response surface $r(x)$. The algorithm requires the user to specify (i) an initial $n$-run design (ii) a function that computes the logarithm of $r(x)$ (in our case, $- \log L(x)$) and (iii) the number of iterations $K$. Each iteration, $n$ new design points are generated and new evaluations of the logarithm of $r(x)$ are done at these points. Thus a total of $Kn$ evaluations of the function are necessary. This algorithm was implemented as the R package ``mined'' in \cite{mined2018}. 

Direct application of the above algorithm to our problem, however, is not possible because even if the true response functions $y_j = f_j(x_1, \ldots, x_p)$ are known for $j=1, \ldots, q$, several points in the initial designs are likely to fall in the unknown infeasible region, not returning any value of the $y_1, \ldots, y_q$ and consequently of $L$. A naive way to avoid this problem is to use a modified response function that only considers the observed responses from the feasible input points, thereby automatically truncating the infeasible points. In the context of Table \ref{tab:data}, this would essentially mean ignoring the lower half of the data matrix, and considering only with the $N_1$ points in the upper half consisting of the feasible points.

Noting that the infeasible region is essentially a ``bad'' region that the MED is trying to avoid, a better strategy is to modify the response function by ``imputing'' the missing responses with values that generates a large value of the loss function. Since to run the MED algorithm we need a function that returns a value of the one dimensional loss for a given input combination, we can simply replace the loss at each infeasible input combination by a value greater than or equal to the largest loss observed from the $N_1$ feasible points in the initial design.

We now illustrate this strategy again using the two-dimensional toy example introduced in Section \ref{sec:intro}. Recall the 20-point initial design shown in Figure \ref{fig:initialtoy}, which generates two infeasible points for which the response values are missing. Among the remaining 18 feasible points, the point $(0.914,0.007)$ near the bottom right corner generates the worst response vector $(0.757,0.085)$ with the largest loss. Therefore, we impute the two missing response vectors with $(0.757,0.085)$ and use all the 20 points as inputs to the MED algorithm. 

\begin{figure}[ht]
\centering
\caption{Contour plot of modified response and MED points generated from initial design} \label{fig:medtoy}
\includegraphics[scale=.6]{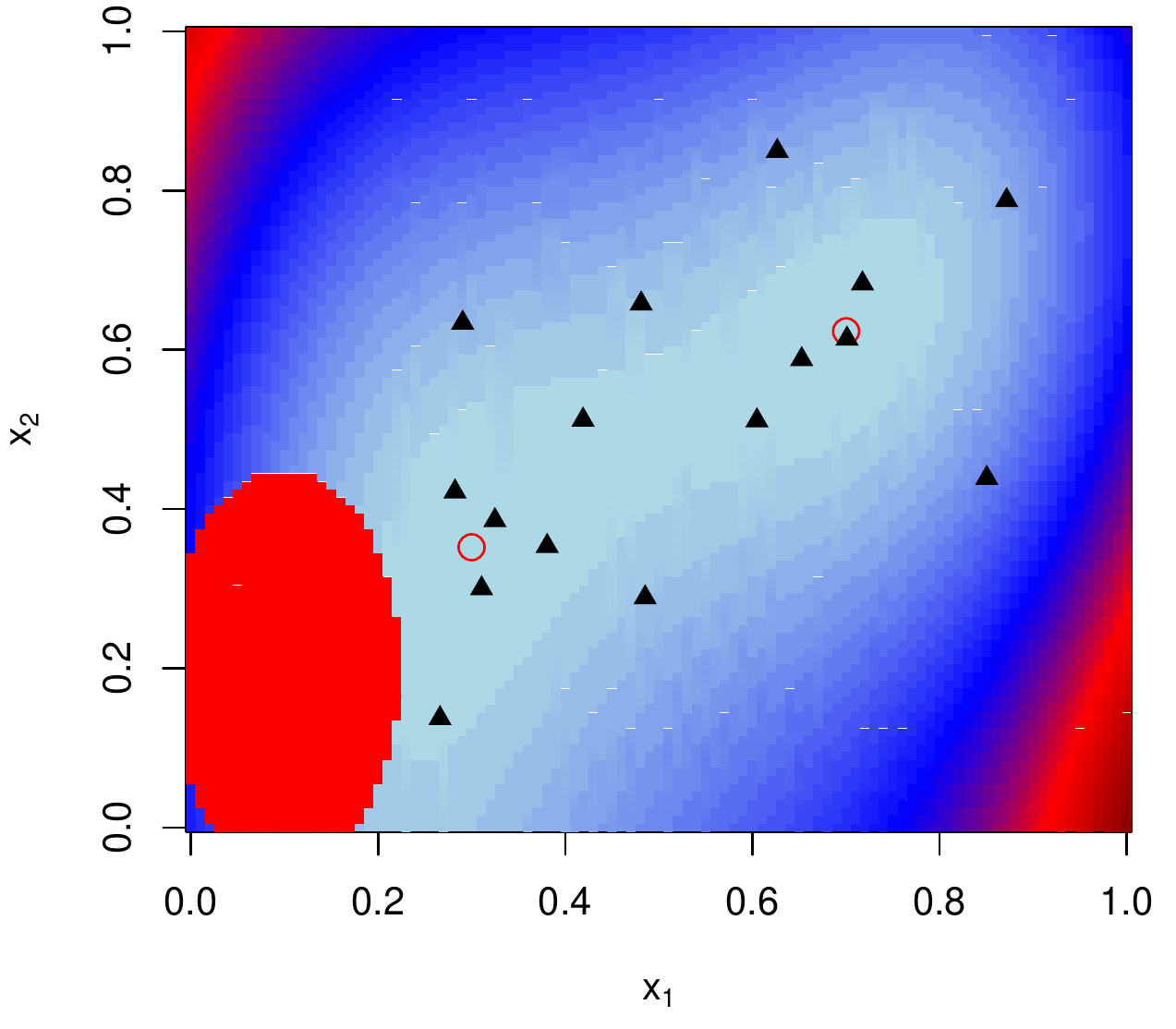} 
\end{figure}

Figure \ref{fig:medtoy} shows the points generated after 4 iterations of the MED algorithm with the contour of the loss function generated from the modified response function shown in the background. It is seen that the MED algorithm beautifully captures the best region, identifies the two optima, and also avoids the infeasible region. Note that one drawback of the R function ``mined'' is that it can produce points outside the input space in an attempt to broaden the search region. In this example, the algorithm ends up generating 16 points in the box $[0,1]^2$ that are shown in Figure \ref{fig:medtoy}.

Although in this particular implementation of MED we end up avoiding generating points in the infeasible region, this is not guaranteed because of the MED algorithm's inherent property of jumping out of good regions and exploring distant regions with uncertainties. Thus, we recommend using the classification model described in Section \ref{ss:classification} to predict the feasibility outcome for each point generated by the MED as the last step. 

\begin{figure}[ht]
\centering
\caption{Performance of the EI algorithm} \label{fig:EI_toy}
\includegraphics[scale=.6]{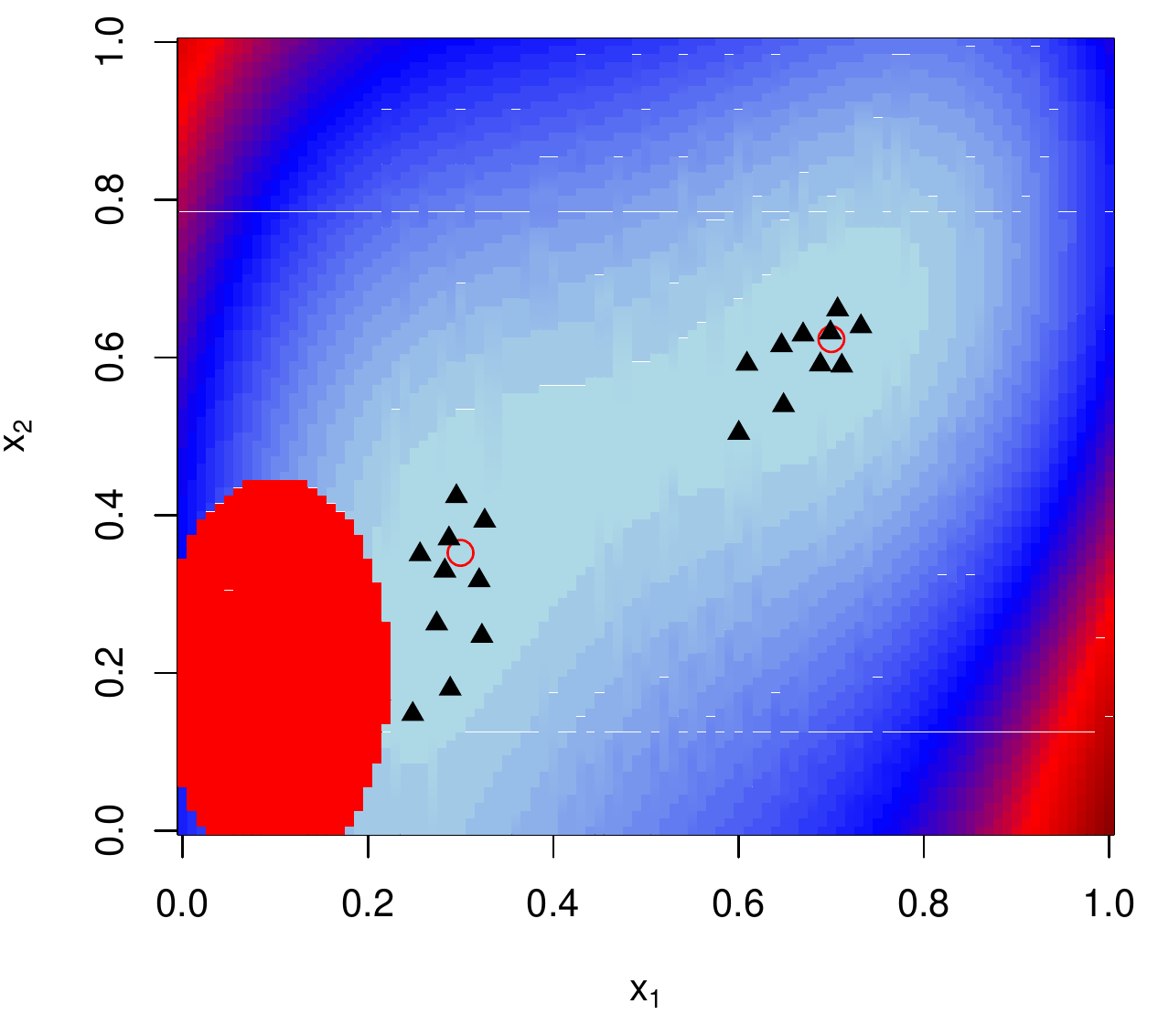} 
\end{figure}

Keeping in mind the motivation behind not adopting a global optimization technique like the BO, as explained in detail in Section \ref{ss:ExpvsOpt}, we compare the performance of the MED with that of the EI algorithm. Figure \ref{fig:EI_toy} shows the performance of the EI algorithm, that uses the same 20-point initial design as in the case of MED and generates 20 new points.

While the EI algorithm identifies two global optima beautifully, all the points cluster around the two global optima in comparison to the proposed algorithm that spreads all the ten points in the best region uniformly, providing more flexibility for ReaxFF users. A measure of this uniformity is the minimum distance between all pairs of points. Larger this measure, the more uniform is the spread of points. This measure is .0013 for the points obtained by the EI algorithm shown in Figure \ref{fig:EI_toy} and .054 for the points obtained by the points obtained by the MED algorithm shown in Figure \ref{fig:medtoy} in the best region. It is also worthwhile to note, that in addition to the ten points in the best region, the MED algorithm also identifies points in the ``next best region'' as indicated by the contour plots. This is a major advantage of the MED algorithm over the EI algorithm, that was also discussed in detail in \cite{MED2015}.

This implementation of the MED algorithm requires $20 \times 4 = 80$ evaluations of the response function. While this is quite reasonable for $p=2$, the number of evaluations necessary will increase with the increase in $p$. If the computation is expensive, one can use a surrogate model or emulator constructed from the initial design and use it as an approximation for the true simulator. We discuss a few strategies to fit such a surrogate model in the following subsection.

\subsection{Surrogate model or emulator of the simulation model} \label{ss:emulator}

Most methods for exploration of deterministic complex response surfaces use Gaussian process models as surrogates. However, while such models are flexible and capable of capturing complex non-linear input-output relationships, even with a moderate number of predictors like 20-25, they become computationally challenging due to identifiability issues associated with model parameters. This is the situation in our motivating problem. For example, the MoS$_2$ ReaxFF system has 45 input parameters and 599 output properties.
 
On the other hand, a naive statistical approach may be to fit individual regression models of $q$ responses on the $p$ inputs, and then predicting the total error based on the individual predictions. However, such an approach is not wise, because there may exist strong correlations among several properties that may not be exploited in the process. Further, any reasonable regression model should at least consider second-order terms, i.e., $p$ square terms and ${p \choose 2}$ pairwise interactions among the parameters. For the MoS$_2$ system, inclusion of the second order terms takes the total number of predictors to 1080, larger than the number of feasible data points obtained in the training data set. This makes linear regression an impossible proposition, although penalized regression methods like Lasso \cite{Tibshirani1996} can be used. Finally, independent prediction of each individual response add up the individual noises or model-fitting errors associated with each fit, resulting in a large prediction error for a utility function that combines individual discrepancies.  

Another popular choice is to fit a multi-response machine learning model such as a deep learning model and use it as a surrogate. In recent years deep learning \cite{DL2015} has emerged as a popular tool for modeling multiple-input multiple-output data. Two main problems associated with fitting deep learning models are (i) they typically require very large training data sets  \cite{Marcus2018} and (ii) tuning these models is not straightforward.

For MoS$_2$ ReaxFF system, we used a combination of simple exploratory techniques for dimension reduction of the response vector and a sequentially fit clustered penalized linear regression that involves first and second order terms of input variables to obtain a surrogate model. Applications of this approach will be demonstrated in Section \ref{sec:application} and details are provided in the Appendix. For Ni-Cr ReaxFF system, which permits quick generation of a very large number of training data points, we fit a deep learning model as the surrogate. 

Let $\hat{f}_j(\mathbf{x})$,  $j=1, \ldots, q$ denote the predictor of $Y_j$  and $\hat{Z}(\mathbf{x})$ denote the predictor of $Z$, the feasibility indicator at input combination $\mathbf{x} = (x_1, \ldots, x_p)$. Then the combined predictor of the loss function $L$ at input combination $\mathbf{x}$ is given by
\begin{equation}
\widehat{L}(\mathbf{x}) = 
\left \{ \begin{array}{cc} 
\sum_{j=1}^q \left\{ \frac{\widehat{f}_j(\mathbf{x}) - T_j}{w_j}\right \}^2, & \hat{Z}(\mathbf{x}) = 1 \\ 
M, & \hat{Z}(\mathbf{x}) = 0,
 \end{array} \right. \label{eq:loss_prediction}
\end{equation}
where $M$ is a number at least as large as the maximum observed loss.

\subsection{Sequential updation and validation} \label{ss:workflow}

Once the MED-step returns a set of promising points based on a surrogate model, it is important to validate them using the actual simulation model. We evaluate each of these points with our classifier to predict their feasibility. Specifically, we predict the response for a point $x^*$ identified by the MED algorithm to lie in the feasible region if the predicted probability $\widehat{\pi}(x^*)$ exceeds a pre-specified threshold $\pi^*$. The points that are predicted to be feasible are then validated by evaluating the true function at those points. Subsequently both the surrogate model and the classifier are updated based on the augmented data.


\section{Simulations} \label{simulation}

We now examine the effectiveness of the proposed approach using simulations from a multiresponse system. We use a slightly modified version of the DTLZ2 function, a popular test function in multi-objective problems \cite{Deb1999, Huband2006}. The advantage of this function is its flexibility - it can be extended to any input and output dimensions $p$ and $q$, and adjusted to create a suitable example for our problem. We first consider the case of $p=q=4$, defining the response functions as
\begin{eqnarray*}
y_1 &=& \{1 + g(\mathbf{x})\} \cos(x_1 3 \pi/2) \cos(x_2 3 \pi/2) \cos (x_3 \pi/2) \\
y_2 &=& \{1 + g(\mathbf{x})\} \cos(x_1 3 \pi/2) \cos(x_2 3 \pi/2) \sin (x_3 \pi/2) \\
y_3 &=& \{1 + g(\mathbf{x})\} \cos(x_1 3\pi/2) \sin(x_2 3 \pi/2), \\
y_4 &=& \{1 + g(\mathbf{x})\} \sin(x_1 3 \pi/2),
\end{eqnarray*}
where $g(\mathbf{x}) = (2x_4 - 0.5)^2$, $\mathbf{x}=(x_1, \ldots, x_4)$. The domain of the functions is $0 \le x_i \le 1$ for $i=1, \ldots, 4$, making the input space is $[0,1]^4$. We also assume that the infeasible region is $D_1 \cap D_2 \cap D_3 \cap D_4$, where $D_1 = \{x_1: 0 \le x_1 \le 0.2\}$, $D_2 = \{x_2: 0 \le x_2 \le 0.2\}$, $D_3= \{x_3: 0 \le x_3 \le 0.2\}$,  $D_4 = \{x_4: 0 \le x_4 \le 0.1\}$. We set the target vector $T = (0.7, 0.7, 0.7, 0.7)$ and the weight vector $w = c(1, 1, 1, 3)$.

\begin{figure}[ht]
\centering
\caption{Contour plot of $L(x_1, x_2, 0.3319, 0.5227)$ and points identified by the MED} \label{fig:DTLZ2_contour}
\includegraphics[scale=.6]{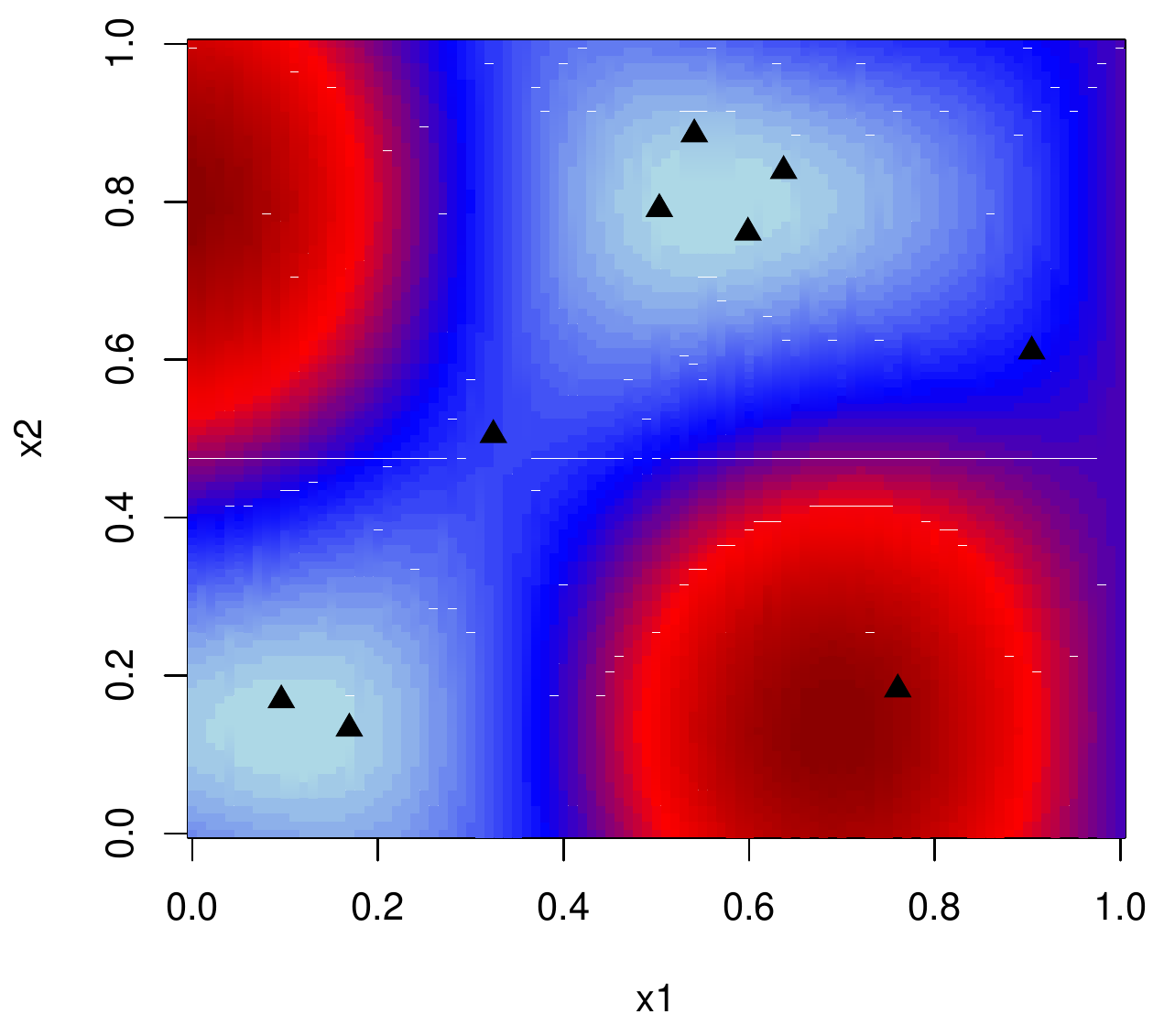} 
\end{figure}

The MED algorithm is applied to this problem with an initial 100-point maximinLHD, adding 100 additional points the proposed algorithm. Figure \ref{fig:DTLZ2_contour} shows the two-dimensional contour plot of the loss function (\ref{eq:lossfn}) against $x_1$ and $x_2$, setting $(x_3, x_4)$ at their best setting (0.3319, 0.5227) identified by the MED in one specific simulation. The figure shows how the MED beautifully identifies the two good regions.  


\begin{figure}[ht]
\centering
\caption{Comparison of proposed algorithm with uniform design: difference of minimum loss (left), difference of median loss (center) and ratio of sd of loss (right)} \label{fig:initialvsMED}
\includegraphics[scale=.32]{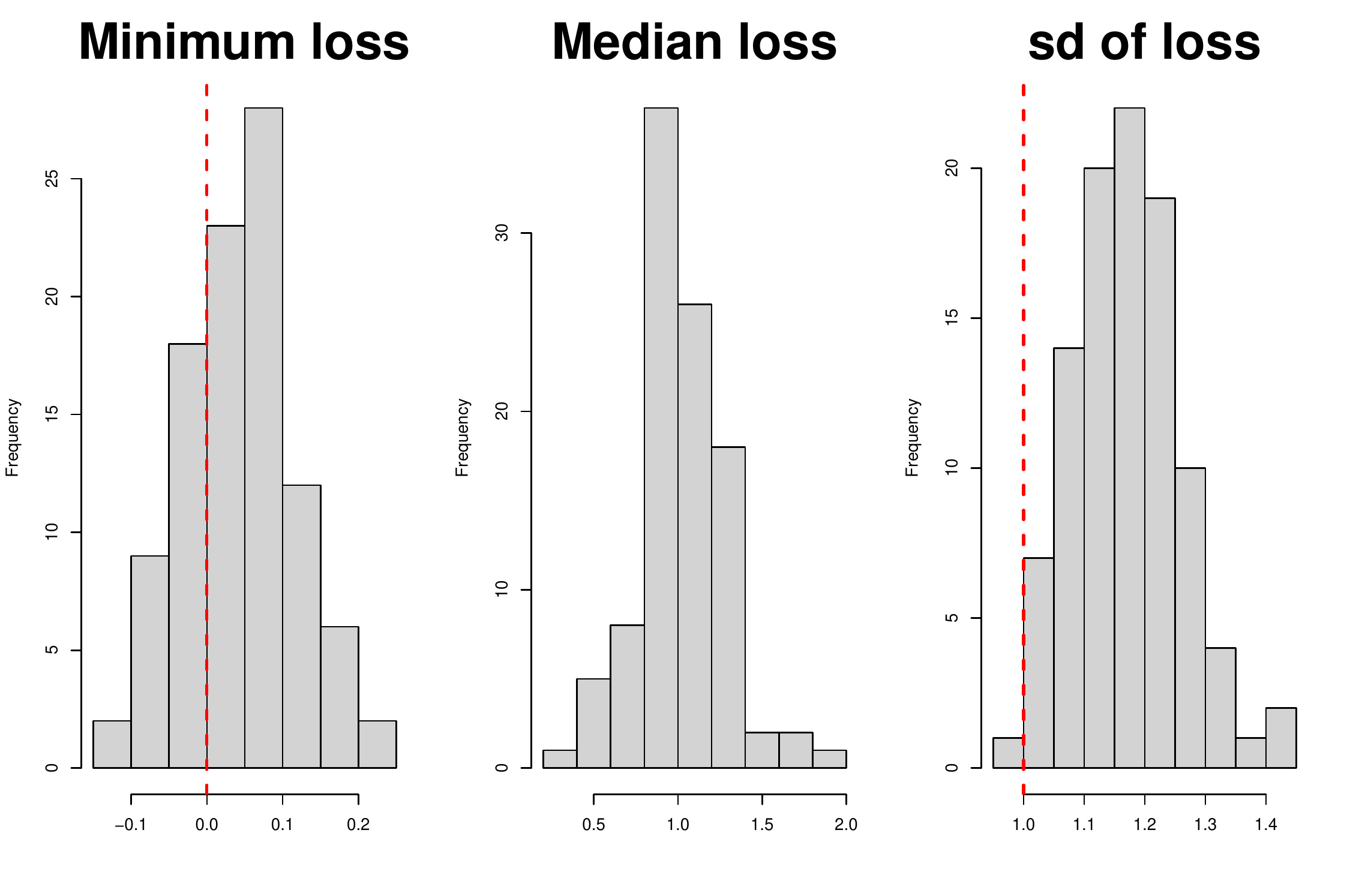} 
\end{figure}

To compare how much improvement is achieved through application of the proposed algorithm in comparison to a random search of the design space, we compare the performance of the 200-run proposed design (consisting of 100 initial maximin LHD points and 100 MED points) with a 200-run uniform design over 100 repetitions. The results are summarized in Figure \ref{fig:initialvsMED}, which compares (i) the difference between the minimum losses identified by the uniform design and the proposed algorithm, (ii) the difference between the median losses identified by the uniform design and the proposed algorithm and (iii) the ratio of the standard deviations of the losses identified by the uniform design and the proposed algorithm. Out of 100 repetitions of the simulation, in 71 cases the MED identified a point with minimum loss lower than the one identified by the uniform design. As expected, in 100\% of the cases the median loss associated with the points identified by the proposed algorithm was less than those identified by the uniform design, and in almost all cases the proposed algorithm had a much smaller variance in the loss.

\begin{figure}[ht]
\centering
\caption{Computation times for MED for different dimensions and initial design sizes} \label{fig:comptime}
\includegraphics[scale=.55]{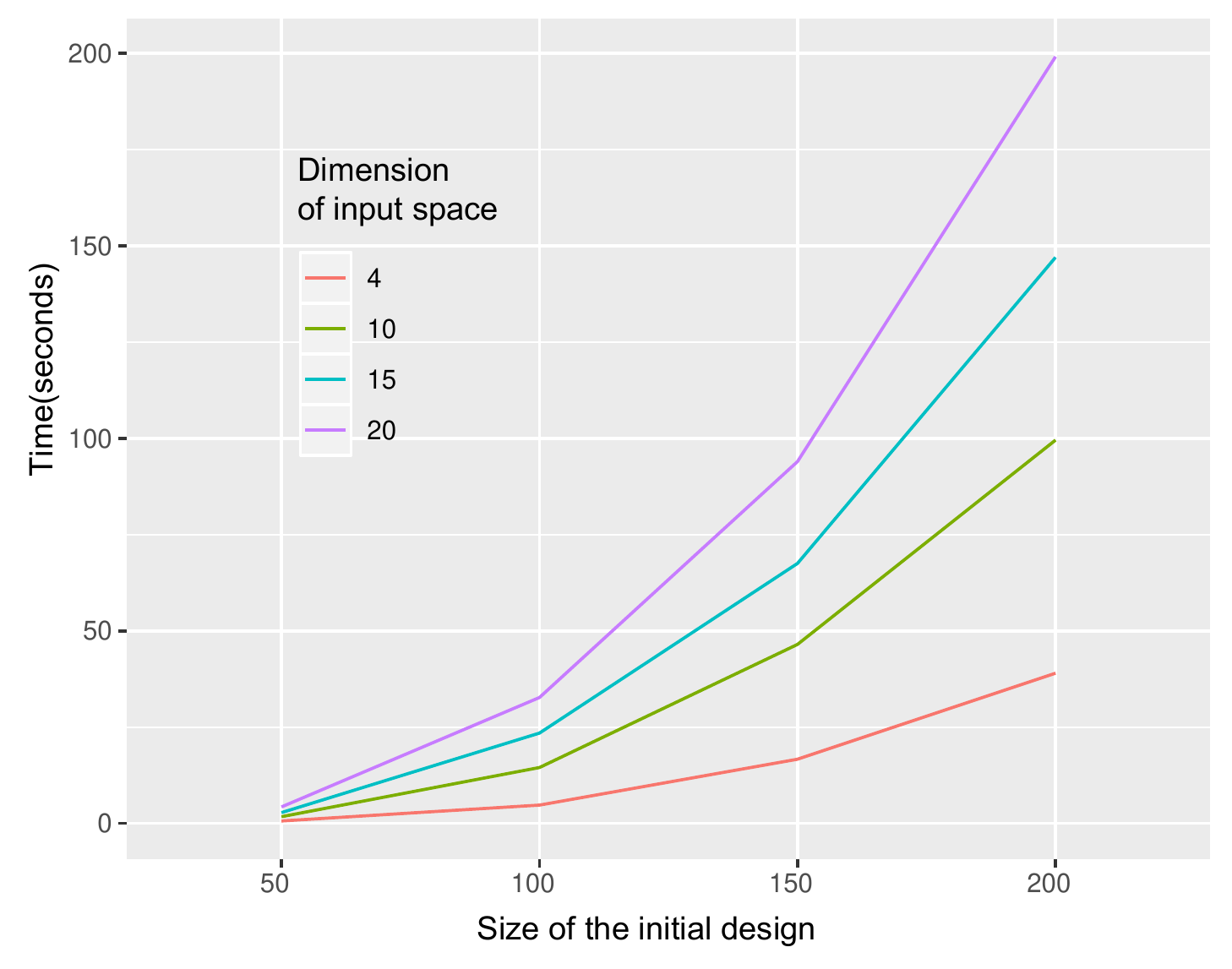} 
\end{figure}

The simulations with the DTLZ2 function described above were also conducted by increasing the input dimension $p$ to 10, 15 and 20. The output dimension $q$ was kept the same as the input dimension. In each case, the size of the initial design $N$ was varied from 50 to 200 in steps of 50, and the number of MED points added was the same as the initial design size. In almost all of these cases, the proposed strategy was able to locate points that were better than the best ones identified by the initial design. The computation times associated with each such simulation settings were also explored, and a representative set of times are shown in Figure \ref{fig:comptime}. As seen from the figure, the computation time taken by the MED algorithm with a 20-dimensional DTZL2 function with 400 points (200 initial and 200 MED points) is approximately 200 seconds.


\section{Applications}\label{sec:application}
We now evaluate the proposed approach with ``Mo-S'' and ``Ni-Cr'' ReaxFF systems. The ``Mo-S system'' is used for materials science and/or physical-chemical systems that are composed of molybdenum (Mo) and sulphur (S) elements, while the ``Ni-Cr system'' is used for systems that are composed of nickel (Ni) and chromium (Cr) elements. The ReaxFF parameters are calibrated against quantum chemical calculations, which are accurate but very expensive. An optimized ReaxFF parameter set can be used to simulate different kinds of materials systems, including bulk crystals, defective structures, and simulation results can be used to calculate defect concentrations, Young’s modulus, and many other properties. These applications depend on how many good points were obtained during optimization, because, as also discussed above, each good parameter set may simulate a different system accurately. In addition, the reference calculations that the parameters are calibrated against are selected in a way to represent materials systems of interest. In this study, Mo-S reference calculations involve two-dimensional surface structures, molecular structures (e.g., MoS$_2$), defect/ad atom structures, some commonly observed reactions, and heat of formation values. The Ni-Cr reference calculations involve bulk crystal structures to calculate equations of state and heat of formation values. The ReaxFF parameters tuned in the Mo-S system include bonding parameters that influence the interactions of bonded atoms, valence angle parameters that influence the vibration of angles, and dihedral parameters that influence the rotation of dihedral angles. The parameters tuned in the Ni-Cr system include bonding and angle parameters and atomic properties such as electronegativity, hardness, Coulombic screening terms for Ni and Cr atoms. 

Different surrogate models were constructed in these two examples, follow by MED algorithm to generate new points.

\subsection{Application to the MoS$_2$ ReaxFF system} \label{sec:MoS2}

\begin{figure}[ht]
\centering
\caption{OMLHD for four pairs of ReaxxFF parameters (only feasible points)} \label{fig:LHD}
\includegraphics[scale=0.60]{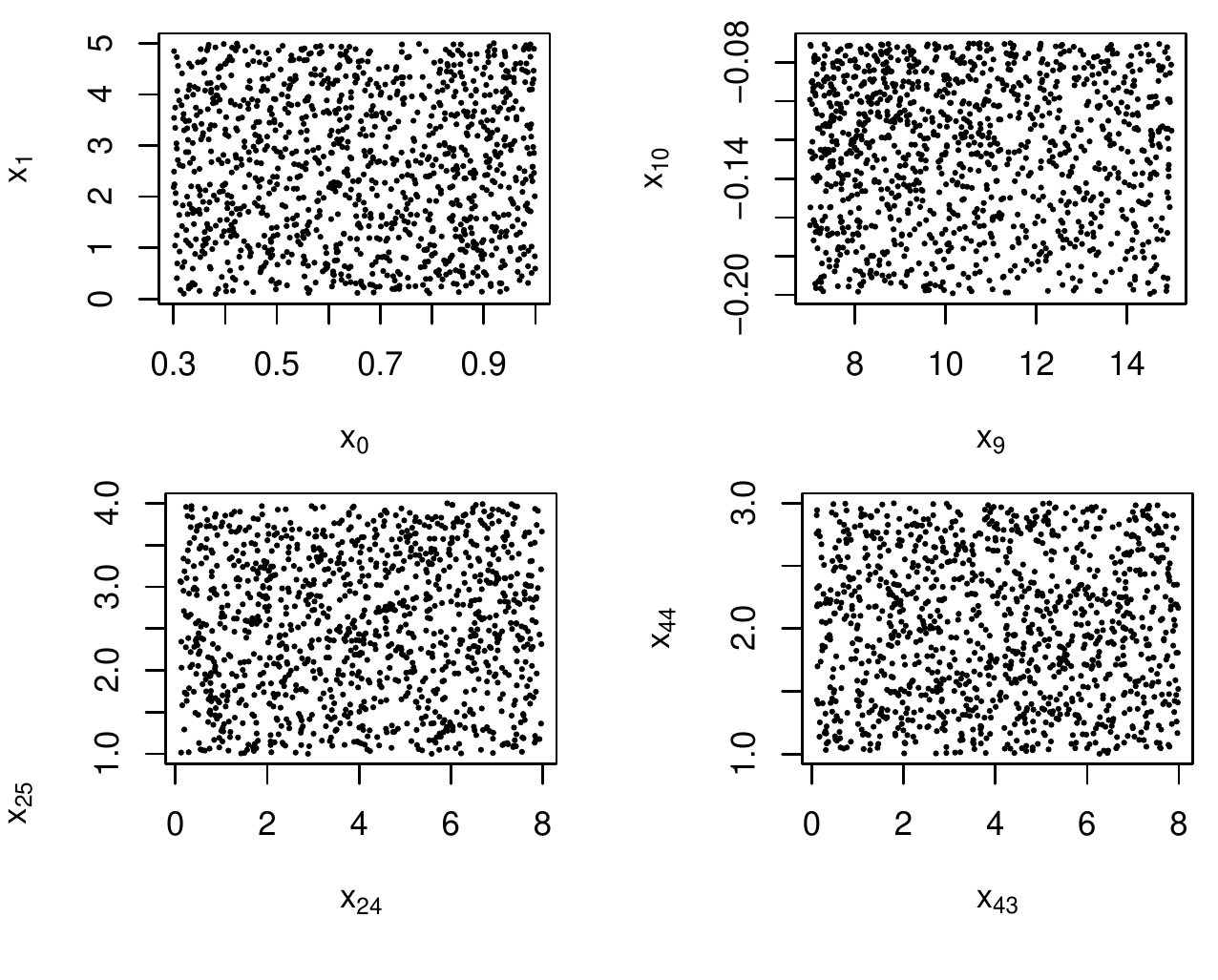} 
\end{figure}

We now demonstrate the proposed approach using the MoS$_2$ ReaxFF system, which has 45 input parameters ($X$s) and 599 material properties or responses denoted by $Y_0, \ldots, Y_{598}$. Using the OMLHD design algorithm described in Section \ref{ss:initial design}, 5000 different combinations of the 45 inputs were obtained. The space-filling property of the OMLHD is illustrated in Figure \ref{fig:LHD} where 2-D scatter plots of four pairs of parameters in the feasible design space are displayed.

Out of the simulation results produced by these 5000 combinations, 3846 were found infeasible and only the remaining 1154 simulations produced results, generating values of 599 different material characteristics. The outcome of each simulation was recorded in the format shown in Table \ref{tab:data} with $N_1 = 1154$, $p = 45$ and $q = 599$. The distribution of the loss function computed from these 1154 points is summarized in Table \ref{tab:summary}.

\begin{table}[htbp]
\centering 
\caption{Distribution of loss function from initial design of size 1154} \label{tab:summary}
\begin{tabular}{c|c|c|c}
min & median & mean & max  \\ \hline
96459 & 2606367 & 17268394  & 848122062 \\ \hline
\end{tabular}
\end{table}

The next step was to fit the classification model to predict the binary outcome variable $Z$ from the $X$ variables. The $5000 \times 46$ data matrix (45 ReaxFF parameters and feasibility outcome $Z$, ignoring the data on responses $Y$'s) was split into a training set consisting of 4000 data points and a testing set consisting of 1000 data points. The models fit with the training data were compared on the basis of their sensitivities and specificities (defined in Section \ref{ss:classification}) computed from the testing data.

A logistic regression classifier that included linear terms of all ReaxFF predictors resulted in a sensitivity of 0.6422 and specificity of 0.9296 based on the testing data. The performance of the random forest classifier, like several other machine learning algorithms, depends on the choice of the tuning parameters. Two of the most important tuning parameters are NTREE (the number of trees to grow) and MTRY (the number of variables that should be selected at a node split). The random forest classifier was fit with the training data with several combinations of NTREE and MTRY, and the performances are summarized in Table \ref{tab:RF_performance}.

\begin{table}[ht]
\centering \scriptsize
\caption{Sensitivity and specificity for random forest classifier with different tuning parameter combinations} \label{tab:RF_performance}
\begin{tabular}{cc|cc}
\multicolumn{2}{c|}{Tuning parameters} & \multicolumn{2}{c}{Performance} \\
NTREE & MTRY & Sensitivity & Specificity \\ \hline
200 & 5 & 0.3088 & 0.9837 \\
200 & 10 & 0.4706 & 0.9611 \\
200 & 15 & 0.5245 & 0.9497 \\
200 & 20 & 0.5147 & 0.9447 \\
200 & 25 & 0.5686 & 0.9372 \\ \hline
400 & 5 & 0.3039 & 0.9925 \\
400 & 10 & 0.4510 & 0.9686 \\
400 & 15 & 0.5245 & 0.9497 \\
400 & 20 & 0.5441 & 0.9410 \\
400 & 25 & 0.5343 & 0.9384 \\ \hline
600 & 5 & 0.2990 & 0.9899 \\
600 & 10 & 0.4461 & 0.9673 \\
600 & 15 & 0.5294 & 0.9472 \\
600 & 20 & 0.5245 & 0.9422 \\
600 & 25 & 0.5343 & 0.9397 \\ \hline
800 & 5 & 0.2843 & 0.9962 \\
800 & 10 & 0.4363 & 0.9673 \\
800 & 15 & 0.5098 & 0.9497 \\
800 & 20 & 0.5245 & 0.9435 \\ 
800 & 25 & 0.5490 & 0.9384 \\ \hline
1000 & 5 & 0.2941 & 0.9925 \\
1000 & 10 & 0.4461 & 0.9673 \\
1000 & 15 & 0.5000 & 0.9510 \\
1000 & 20 & 0.5392 & 0.9410 \\
1000 & 25 & 0.5441 & 0.9384 \\
\end{tabular}
\end{table}

It appears from Table \ref{tab:RF_performance} that in terms of predicting the feasibility of ReaxFF parameter combinations, the logistic regression based classifier outperforms the random forest classifier with respect to sensitivity, and is marginally inferior with respect to specificity. Considering the role of this classifier in predicting whether a promising new combination of ReaxFF parameters should be added to the exploration space, a higher true positive rate is possibly more important than achieving a higher true negative rate. This is because, a positive outcome (feasible combination) leads to a successful simulation and generates values of material characteristics, whereas a negative outcome or infeasible point does not add anything to the existing knowledge. From this perspective, the logistic regression based classifier was chosen over the random forest classifier.

\begin{figure}[ht]
\centering
\caption{\small Plot of means and sds of 599 responses obtained from 1154 data points; the $+$ sign represents 37 shortlisted responses} \label{fig:mean-sd}
\includegraphics[scale=.45]{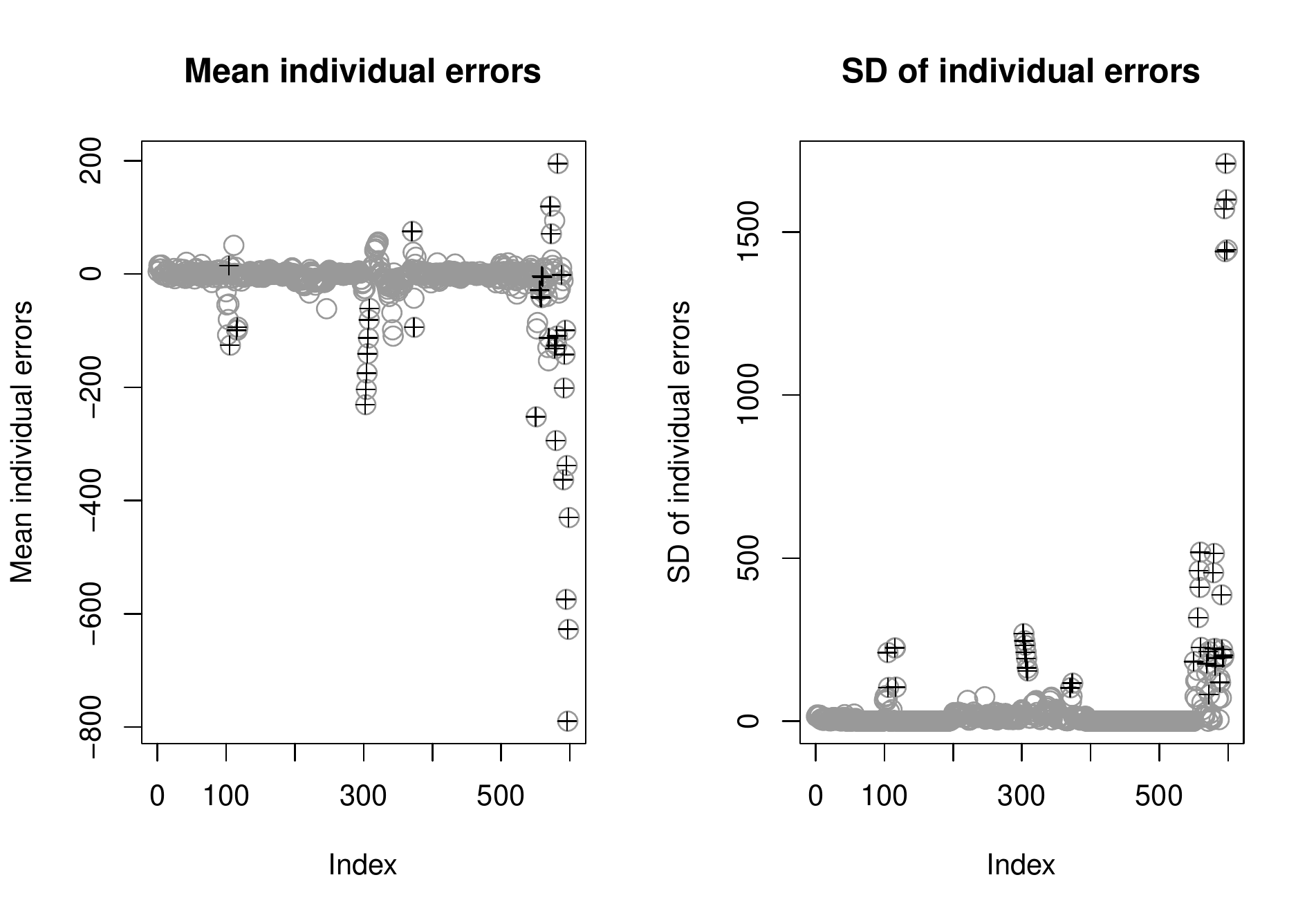} 
\end{figure}

As a pre-cursor to fitting the surrogate model to predict $Y$'s from the $X$'s, an exploratory analysis, following the guidelines provided in Appendix \ref{ss:exploratory}, was conducted with the $1154\times 599$ matrix of $Y$s by summarizing the scaled individual errors (SIE)  $E_{ij} = (Y_{ij} - T_j)/w_j$ for $i=1, \ldots, 1154$ and $j = 1, \ldots, 599$ in an attempt to identify the ones that have major contributions to the total discrepancy. A graphical summary of the means and standard deviations of SIEs of the 599 responses (calculated from 1154 observations) is shown in Figure \ref{fig:mean-sd}.

Figure \ref{fig:mean-sd} revealed a very interesting aspect - several (in fact 190 out of 599) of the responses appeared to have negligibly small standard deviations in the generated data, meaning that they remained more or less constant over the 1154 feasible ReaxFF parameter combinations in the initial design. Clearly prediction models involving such properties are meaningless. Interestingly, most of these 190 properties also had their means very close to the gold standards, meaning they could not be improved further. On the other extreme, there were some responses that varied widely across the parameter settings and were therefore potentially interesting candidates for model fitting. As proposed in Appendix \ref{ss:exploratory}, using the measure $P(\mathcal{J}^*)$ given by (\ref{eq:percentage}), 37 properties were identified as the top contributors to the total squared SIE $\sum_{i=1}^{1154} \sum_{j=1}^{599} E^2_{ij}$. These 37 properties contributed to 96.7\% of the total SIE in the generated data, i.e., $P(\mathcal{J}^*) = 0.967$ where $\mathcal{J}^*$ represents the set of indices of these properties. These 37 properties are represented by $+$ signs in both panels of Figure \ref{fig:mean-sd}. 

\begin{figure}[htbp]
\centering
\caption{\small Correlation heatmap among 37 log absolute weighted individual errors (red color indicates
high correlation)} \label{fig:corr}
\includegraphics[scale=.35, angle=-90]{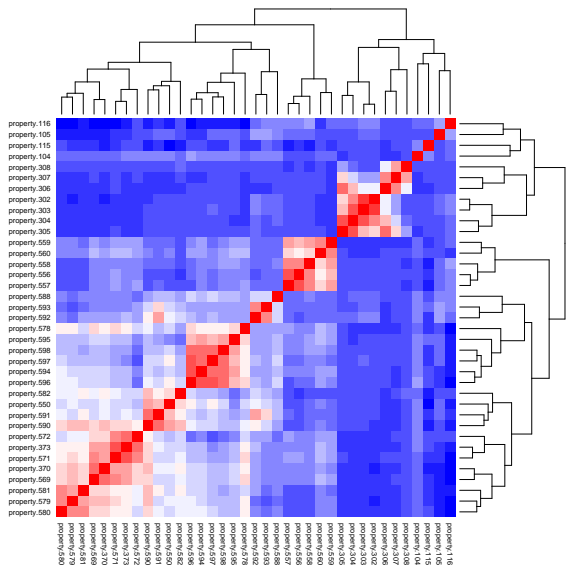} 
\end{figure}

The 37 shortlisted properties were grouped into five clusters on the basis of correlations among the transformed responses $U_j = \log |(Y_j - T_j)|/w_j$, i.e, the logarithms of absolute individual differences from the target values. The logic behind transforming the responses is explained in Appendix \ref{ss:model-fitting}. Figure \ref{fig:corr} shows the heatmap of correlations. 

Finally, the predictor of the loss function $\widehat{L}(\mathbf{x})$ at a new input combination $\mathbf{x}$ was obtained along the lines of (\ref{eq:loss_prediction}) and (\ref{eq:ReaxFF_loss_pred}). After running the MED algorithm with this predicted loss function, we were able to identify several promising points. The feasibility of each of these points was evaluated using the logistic regression classifier, and the points for which the feasibility probability was found to be 0.5 or larger were validated with the actual ReaxFF simulation. The resultant points were used to update the classifier and surrogate model, as described in Section \ref{ss:workflow}. After repeating this cycle a few times, several points were found to produce excellent results with the actual ReaxFF simulation, and in particular, one point was found with a loss of approximately 80,000, providing a substantial (17\%) improvement over the best point obtained from the initial 1154 points.

\subsection{Application to the Ni-Cr ReaxFF system} \label{sec:NiCr}
We now illustrate the proposed approach using another ReaxFF system Ni-Cr, which has 16 input parameters $(X s)$ and 90 material properties denoted by $Y_0, \ldots, Y_{89}$. The main difference in this application from the previous example of the $\text{MoS}_2$ system was the surrogate modeling approach. Unlike the cluster-based sequential penalized regression model used  for exploring the $\text{MoS}_2$ system, we decided to fit a deep learning (DL) model as a surrogate. Fitting a DL model typically entails generating a large number of training data points. Therefore, using the OMLHD design algorithm described earlier, 79635 different combinations of the 16 inputs were obtained, and ReaxFF simulations were conducted with these combinations. Out of these 79635 combinations, 4999 were found infeasible. The rest produced meaningful values of the responses. The distribution of the loss function obtained from these points is summarized in Table \ref{tab:summary_NiCr}.


\begin{table}[htbp]
\centering 
\caption{Distribution of loss function from Ni-Cr initial design} \label{tab:summary_NiCr}
\begin{tabular}{c|c|c|c}
min & median & mean & max  \\ \hline
2812 & 2416505 & 63081433  & $1.33 \times 10^{12}$ \\ \hline
\end{tabular}
\end{table}  

The data obtained from the feasible region were split into two parts - a training set comprising $80\%$ of the points, and a testing set consisting of the remaining 20\%. The training data were used to fit the DL model with two hidden layers and 90 nodes and the model was tested using mean absolute error (MAE) as the measure of prediction accuracy. Details on fitting the DL algorithm can be found in \cite{Sengul2020}.

The trained DL model permitted prediction of material properties $\widehat{\mathbf y}$ at any new input combination $\mathbf{x}$. Consequently, the predictor of $\widehat{L}(\mathbf{x})$ was obtained along the lines of (2) and (4). The initial 200-run design for MED was chosen from a region centered around the parameter combination with the smallest loss obtained from the initial design points. After running the MED algorithm with the surrogate loss function based on the DL model, several promising points were obtained. Following a workflow described in Section \ref{ss:workflow}, and similar to the MoS$_2$ system, validation experiments were conducted with the actual ReaxFF simulation model and the training models were updated. Eventually, several points with losses smaller than 2812 were identified. The best point with a loss 2316.17 provided a substantial ($18\%$) improvement over the best point obtained from the initial 79635 points. Our approach produced a significant improvement over the conventional method, which was not able to find input points with loss lower than 5614.26. In addition, it also stood out in terms of computation time by producing several good parameter combinations, in addition to the optimum, in about two minutes compared to two weeks taken by an experienced force field developer using the conventional method. This was an artifact of the MED-based algorithm.



\section{Concluding remarks} \label{sec:conclusion}

In this article, we have proposed a framework for finding good points in an input space that produce a vector of responses close to their target values, where the ``goodness'' or ``closeness'' is defined in terms of a one-dimensional loss function. This problem is different from traditional  global optimization because of the possible existence of several good regions with almost similar values of the loss function, and there are large unknown input regions that are infeasible in the sense that they do not produce any output. The responses are assumed to be deterministic, but can be extended to accommodate noise. The problem is motivated by and applied to the ReaxFF optimization systems in physics, and provide initial results that are encouraging. Two key elements of the proposed approach are classifying the input space into feasible and infeasible regions and sampling points from the good regions using a method called minimum energy designs.

Another popular approach to define ``goodness'' of points in multiresponse systems is to find solutions that are Pareto optimal \cite{Cook2011, Cook2014a, Cook2014b,Chen2018}, i.e., solutions that cannot be improved so as to make any one response closer to its target value without making at least another response move farther away from its target value. However, finding Pareto optimal solutions with unknown constraints, and input and output dimensions as high as what is encountered in a typical ReaxFF problem is rarely addressed. This can be an interesting direction for future research.

\section{Appendix} \label{sec:appendix}

\subsection{Exploratory analysis, dimension reduction and clustering of responses} \label{ss:exploratory}

Recall that our objective is to obtain input combinations that produce low discrepancies of responses from respective target values, i.e., $Y_j - T_j$ for $j=1, \ldots, p$. We first take a close look at the $N_1 \times q$ matrix of scaled individual errors (SIE)  $E_{ij} = (Y_{ij} - T_j)/w_j$, for $i=1, \ldots, N_1$ and $j = 1, \ldots, q$, where $w_j$'s denote weights, in an attempt to identify the ones that have major contributions to the total discrepancy. Plots of summary statistics of the SIE's, i.e., their sample means $\bar{E}_j = \sum_{i=1}^{N_1} E_{ij}/N_1$, sample standard deviations $s_j =  \sqrt{ \sum_{j=1}^{N_1}(E_{ij} - \bar{E}_j)^2 / (N_1 - 1)}$ for $j = 1, \ldots, q$ or sample quantiles provide useful information on the distributions of SIEs for each response. 

Based on such exploratory analysis, one can choose a subset of output characteristics that contributes to at least an aimed proportion of the total squared SIE $\sum_i \sum_{j} E_{ij}^2$ as a criterion. That is, we can choose a subset $\mathcal{J}^*$ of the set of indices $\{1, \ldots, q\}$ satisfying $P(\mathcal{J}^*) \ge \delta$, where
\begin{equation}
P(\mathcal{J}^*) = \frac{ \sum_{i=1}^N \sum_{j \in \mathcal{J}^*} E^2_{ij} }{\sum_{i=1}^N \sum_{j=1}^{q} E^2_{ij} }  \label{eq:percentage}
\end{equation}
represents the proportion of total squared SIE from all responses explained by the responses $Y_j$ for $j \in \mathcal{J}^*$ in the data generated and $0 < \delta < 1$ is a chosen threshold, which could be 0.80 or 0.95, for example.

\subsection{A cluster-based sequential penalized regression model} \label{ss:model-fitting}

We now consider prediction of the responses $Y_j$ for $j \in \mathcal{J}^*$, where $\mathcal{J}^*$ is the set of responses identified in the previous section. Let $q^* \le q$ denote the cardinality of $\mathcal{J}^*$. A naive strategy is to fit  independent regression models of each of the $q^*$ responses on the input variables $X$. However, such a strategy does not take into account potential correlations present within the properties. We propose fitting a cluster-based penalized regression model to address this problem. Instead of modeling the responses $Y$, specially for the ReaxFF application, we propose modeling the transformed responses (TR) $U = \log|Y - T|/w$ where $T$ is the target value for $Y$ and $w$ is the weight defined earlier. This transformation is justified by the fact that it is the absolute difference between the response and target value that needs to be made small. Further, in the context of the ReaxFF problem, for most of the responses the distribution of the absolute individual error $|(Y-T)/w|$ appeared to have moderate to heavy skewness, that could be corrected using the log transformation. 

The first task is to cluster the $q^*$ TR variables $U$ into $C$ clusters based on their correlation matrix. This can be done by using any clustering package in a standard statistical computing. Let $C$ denote the number of clusters and $K_1, \ldots, K_C$ the number of TR variables in clusters $1, \ldots, C$ respectively.

Let $U_{\ell 1}, \ldots, U_{\ell K_l}$ denote the $K_{\ell}$ TRs for cluster $\ell \in \{1, \ldots, C\}$. We use the following algorithm to fit a set of predictor models for $U_{\ell 1}, \ldots, U_{\ell K_l}$.
\begin{enumerate}
\item[(i)] For each TR $U$ in the cluster, using a Lasso regression \cite{Tibshirani1996}, select significant predictors in the penalized linear model of $U$ on all $2p + {p \choose 2}$ first and second order terms, i.e., $X_h$, $X_h^2$ and $X_h X_{h^{\prime}}$ for $h, h^{\prime} = 1, \ldots, p, h \ne h^{\prime}$ . 
\item[(ii)]From these $K_{\ell}$ models, select the one with the maximum predictive power (determined using out-of-sample mean-squared error, cross validation methods or adjusted $R^2$). This becomes the baseline model of the cluster, interpreted as the one with maximum predictive power solely based on the input variables ${\mathbf X}$. Let $U^{*}_{\ell 1}$ denote the TR in the baseline model, and we call $U^{*}_{\ell 1}$ the baseline TR in cluster $\ell$. Let $\widehat{U}^{*}_{\ell 1} = f^{*}_{\ell 1}(\mathbf{x})$ represent the baseline model.
\item[(iii)] Now consider the prediction of the remaining $K_{\ell}-1$ TRs in this cluster. Pick the TR, say $U^{*}_{\ell 2}$, that can be best predicted using the predictors chosen by Lasso in step (i), and the baseline TR $U^{*}_{\ell 1}$ already identified in step (ii). Then update the model for  $U^{*}_{\ell 2}$ by including the baseline TR $U^{*}_{\ell 1}$ as a predictor in addition to the $X$ terms if it satisfies a pre-specified inclusion rule (like achieving a threshold improvement in out-of-sample error or adjusted $R^2$). Let $\widehat{U}^{*}_{\ell 2} = f^{*}_{\ell 2}(\mathbf{x}, U^{*}_{\ell 1})$ denote this model.
\item[(iv)] Repeat step (iii) sequentially for the remaining TRs in the cluster, and include TRs from the previous steps to predict them in addition to the $X$'s if the inclusion rule is satisfied.
\end{enumerate}

The above procedure is repeated for each cluster. Thus we have, for each cluster, a collection of models that predict each TR in that cluster. Let $\widehat{U}_{\ell k}(\mathbf{x})$ denote the predicted TR for the $k$th response of cluster $\ell$. Then the surrogate model for the loss function at input combination $\mathbf{x}$ is
\begin{equation}
\widehat{L}(\mathbf{x}) = \left\{ \begin{array}{cc} 
\sum_{\ell=1}^C \sum_{k=1}^{K_{\ell}} \left\{ \exp \left( \widehat{U}_{\ell k}(\mathbf{x}) \right) \right\}^2, & \widehat{Z}(\mathbf{x}) = 1, \\
M & \widehat{Z}(\mathbf{x}) = 0, 
\end{array} \right. \label{eq:ReaxFF_loss_pred}
\end{equation}
where $M$ is a very large number and $ \widehat{Z}(\mathbf{x})$ the predicted binary response outcome from the classification model.

\section*{Acknowledgment}
The research was supported by NSF EAGER Grant No. DMR 1842952, DMR 1942922 and MRI 1626251.

\bibliographystyle{apalike}
\bibliography{ReaxFF_references}

\end{document}